\documentclass{article}

\usepackage{microtype}
\usepackage{graphicx}
\usepackage{booktabs} 
\usepackage{hyperref}
\usepackage{enumitem}
\usepackage{microtype}
\usepackage{array}
\usepackage{tabularx}
\usepackage{float}
\usepackage{amssymb}
\usepackage{seqsplit}
\usepackage{cuted}
\usepackage{caption}
\usepackage{graphicx}    
\usepackage{subcaption}

\usepackage{xurl}
\newcommand{\code}[1]{\nolinkurl{#1}}
\newcolumntype{Y}{>{\raggedright\arraybackslash}X}
\usepackage{tikz}

\usepackage{booktabs}
\usepackage{tabularx}
\usepackage{array}
\usepackage{makecell}
\usepackage{amssymb}

\newcommand{\circled}[1]{%
  \tikz[baseline=(char.base)]{
    \node[shape=circle,draw,inner sep=1pt,font=\bfseries] (char) {#1};
  }%
}

\usepackage[accepted]{mlsys2025}

\newif\ifshowcomments
\showcommentstrue      
\ifshowcomments
  \usepackage{xcolor}
  \newcommand{\mm}[1]{\textcolor{magenta}{Maleeha: #1}}
  \newcommand{\igc}[1]{\textcolor{blue}{IG Comment: #1}}
  \newcommand{\igt}[1]{\textcolor{green}{IG Text: #1}}
  \newcommand{\dv}[1]{\textcolor{red}{DV: #1}}
\else
  \newcommand{\mm}[1]{}
  \newcommand{\igc}[1]{}
  \newcommand{\igt}[1]{}
  \newcommand{\dv}[1]{}
\fi

\usepackage{graphicx}

\newcommand{\para}[1]{\vspace{6pt}\noindent\textbf{#1}}
\newcommand{\name}{Trail}
\newcommand{\expansion}{{\it \underline{T}yped \underline{R}egistry-\underline{A}nchored \underline{I}nference for \underline{L}anguage commands in Smart Homes}}

\mlsystitlerunning{\name{}: Accurate Local Smart Home Control Using Decision Models}

\usepackage{listings}
\lstdefinestyle{prompt}{basicstyle=\ttfamily\footnotesize, breaklines=true, columns=fullflexible,
  frame=single, framesep=4pt, xleftmargin=4pt, xrightmargin=4pt, keepspaces=true}

\begin{document}

\twocolumn[
\mlsystitle{\name{}: Accurate Local Smart Home Control Using Decision Models}


\begin{mlsysauthorlist}
\mlsysauthor{Maleeha Masood}{to}
\mlsysauthor{Deepak Vasisht}{to}
\mlsysauthor{Indranil Gupta}{to}
\end{mlsysauthorlist}

\mlsysaffiliation{to}{University of Illinois Urbana-Champaign}

\mlsyscorrespondingauthor{Maleeha Masood}{maleeha2@illinois.edu}

\printAffiliationsAndNotice{}

\vskip 0.3in

\begin{abstract}
Natural-language driven smart home control remains challenging: cloud-hosted Large Language Models (LLMs) raise privacy concerns, while local Small Language Models (SLMs) can generate invalid actions or misinterpret requests. We introduce \name{} (\expansion{}), a fully local framework that replaces open-ended command generation with hierarchical typed decisions. \name{} exploits the natural structure of a home's action space, selecting a room, device, and operation from candidates derived from the runtime device registry. This ensures valid actions while enabling confidence-based abstention when requests cannot be reliably resolved. We also design a structured memory module that uses typed decisions to support user preferences, device references, and multi-turn interactions. To evaluate \name{}, we extend HomeBench~\cite{homebench} and generate three datasets: \emph{Trek-Base}, which tests natural-language variation; \emph{Trek-Robust}, which evaluates robustness to changes in the home and its description; and \emph{Trek-Memory}, which assesses conversational memory across multi-turn interactions. \name{} achieves 85.3\% accuracy on Trek-Robust versus 69.8\% for a finetuned Qwen3.5-4B baseline, while remaining competitive on Trek-Base requests. With memory, \name{} resolves 75.6\% of the requests that depend on earlier turns, compared with 42.5\% for the finetuned baseline given conversation history.
\end{abstract}
]

\section{Introduction}
Adapting smart homes to user preferences has been a longstanding research area \cite{mozer1998neural, 5196706, sasha, vcullm, adahome, domux}. With recent advances in Large Language Models (LLMs), natural-language interfaces allow for more conversational smart home control. Accurately interpreting  such requests is especially important because incorrect actions have physical consequences such as leaving an oven running unattended or unlocking the front door at night. Today, the interpretation of user requests often relies on cloud-hosted LLMs, requiring private user data to be sent to third parties for inference and risking service disruptions during network outages. Small Language Models (SLMs) offer a promising alternative through local inference, but they struggle with complex tasks and generalization~\cite{ibm_slm}.

We present \name{} -- \expansion{} -- an entirely local smart home control framework that is accurate, completely private, and is always available. \name{} leverages the new generation of Decision Models (aka System One models), such as Jev \cite{almeida2026systemone} and Microsoft-Decision-1 \cite{srivastava2026microsoftdecision1}. Unlike traditional language models that generate free-form text, Decision Models score predefined choices and return their associated confidence scores. Their non-generative inference enables fast, structured decision-making, making them promising for interactive local smart home control.

\name{} formulates smart home control as a selection problem using typed decisions. Each typed decision presents a question and a set of predefined choices, which a Decision Model scores and selects from. At runtime, these choices are derived from a home's device registry, which defines the available devices, their supported operations, and their parameter constraints. This allows \name{} to adapt to new devices without retraining and refuse requests that cannot be reliably resolved, rather than guessing --- such a capability of ``I don't know how to do it'' is a missing feature from today's LLMs, which do not stop trying even when the task is impossible. Furthermore, \name{} runs entirely locally, preserving user privacy and availability. In designing \name{}, we make the following key contributions.

\para{Smart Home Control as Hierarchical Decisions.} 
Our key insight is that a smart home's action space has a natural hierarchical structure: resolving a user request involves first identifying the relevant room, then selecting a device within that room, and finally determining the operation to perform on that device. For example, given the request ``dim the living room light to 40\%'', \name{} first selects the living room, then the light within that room, and finally the brightness operation of the selected light. Each decision considers only the candidates consistent with previous selections, reducing the number of choices the model must evaluate at each step.

\para{Integrating User Preferences and References.} 
We design a structured memory module that allows \name{} to remember and recall user {\it preferences} and {\it references}. (1) \emph{Preferences} store a preferred value for a device's setting, optionally tied to a situation (e.g., reading-lamp brightness 40 when reading). (2) \emph{References}  map a user's own words to a device (e.g., ``the front light'' to \texttt{foyer.light}), including the device a user has just operated so that ``turn it off'' can refer back to it. Memory is managed using typed decisions: they decide what to store, link each stored record to a device in the home's registry, and retrieve the relevant records when a later request needs them. This enables personalized, context-aware control without placing conversation history in the SLM's prompt.

\para{New Smart Home Datasets.} 
To address the paucity of natural-language datasets for smart homes, we extend HomeBench~\cite{homebench} to generate 3 smart home datasets. First, we introduce \emph{Trek-Base}, which tests natural-language variations in user requests, including implicit device references and goal-oriented commands. Second, we develop \emph{Trek-Robust}, a stress test that evaluates robustness to changes in the home and its description. Third, we build \emph{Trek-Memory}, that converts Trek-Base into 100 multi-turn conversations, one per home, with device states that carry over between turns and inserted requests that require memory, such as user-defined device nicknames and corrections.

We implement \name{} using JevK5~\cite{openjev}, an open-source Decision Model built on Qwen3.5-4B. We evaluate \name{} on Trek-Base, Trek-Robust and Trek-Memory against direct-generation, retrieval, agentic and slot-filling baselines. \name{} stays within 4 points of the strongest baseline on Trek-Base, makes 2.1$\times$ fewer errors on Trek-Robust, and resolves 2.1$\times$ more of the requests that require memory.
\begin{table*}[!t]
\centering
\small
\renewcommand{\arraystretch}{1.15}
\setlength{\tabcolsep}{5pt}

\begin{tabularx}{\textwidth}{%
    >{\centering\arraybackslash}p{0.01\textwidth}
    >{\raggedright\arraybackslash}p{0.2\textwidth}
    >{\raggedright\arraybackslash}p{0.18\textwidth}
    *{3}{>{\raggedright\arraybackslash}X}
}
\toprule
& \textbf{User Requests}
& \textbf{Correct Output}
& \textbf{Qwen2.5 1.5B}
& \textbf{Qwen2.5 3B}
& \textbf{Qwen2.5 7B} \\
\midrule

\textbf{\circled{1}}
& dim the balcony light to 50\%
& \code{Ba.l.br(50)}
& \checkmark
& \checkmark
& \checkmark
\\

\textbf{\circled{2}}
& make it a bit warmer \textit{(21°C now)}
& \code{L.th.temp(22)}
& \code{L.th.temp(24)}
& \checkmark
& \checkmark
\\

\textbf{\circled{3}}
& turn on the bedroom AC \textit{(no bedroom AC)}
& \code{error_input}
& \code{L.th.temp(24)}
& \code{L.th.temp(26)}
& \code{B.AC.on()}
\\

\textbf{\circled{4}}
& turn off the light
& \code{error_input}
& \code{L.l.off()}
& \code{L.l.off()}
& \code{L.l.off()}
\\

\textbf{\circled{5}}
& set the balcony light to 150\%
& \code{error_input}
& \code{Ba.l.br(150)}
& \code{Ba.l:br(120)}
& \code{Ba.l.br(100)}
\\


\textbf{\circled{6}}
& turn on the living room light at 40 and make it 2 degrees cooler \textit{(21°C now)}
& \code{L.l.br(40), L.th.temp(19)}
& \code{L.l.on() L.th.temp(20)}
& \code{L.l.on(), L.th.temp(18)}
& \code{L.l.on(), L.th.temp(19)}
\\

\textbf{\circled{7}}
& good night, secure the front door
& \code{F.lock.lock()}
& \code{lock(F)}
& \code{F.lock()}
& \code{F.lock.unlock(), F.lock.lock()}
\\

\bottomrule
\end{tabularx}

\caption{\centering Qwen2.5 outputs across different model sizes for requests to the smart home configuration in \S\ref{sec:slmfailure} -- prompt used in Appendix \ref{app:tableprompt}. For readability, \code{Ba} is balcony, \code{B} is bedroom, \code{L} is living\_room, \code{F} is front\_door, \code{l} is light, \code{th} is thermostat, \code{br} is set\_brightness, \code{temp} is set\_temperature, and \checkmark is correct. \name{} gets all correct except \circled{2}, where the temperature is set to 26°C.}
\label{tab:model-outputs}
\end{table*}

\section{Background: Decision Models and Typed Decisions}

Decision Models \cite{almeida2026systemone, deußer2026evaluatingbenchmarkingmodeljev, rafe2026benchmarkingdecisionmodelstrained, sharma2026typedbenchbenchmarkcalibrationframing} are non-generative LLMs that make fast, structured decisions over predefined outputs using typed decisions. Instead of generating output tokens, a Decision Model scores the candidates of the typed decision and returns the choices with their associated confidence score. 

A typed decision restricts the model's output to a predefined type and set of valid choices, such as selecting a device from those available in a home. For example, given the user command \emph{``Turn on the kitchen lights''}, a smart home controller can define a typed decision over the set of available devices, such as \code{light, switch, coffee maker, fridge}. The model scores these candidates and selects \code{light}, rather than generating a free-formed device name. This ensures that the selected device belongs to the set of valid options.

Decision Models are particularly well suited to smart home control, where the available devices are known, but may vary across homes and change over time. Typed decisions allow the model to select from these at runtime. \name{} uses JevK5 \cite{openjev} to provide this typed decision interface. To the best of our knowledge, we are the first to use Decision Models for local smart home control.

\section{Motivation}
\label{sec:slmfailure}

We describe two failures in local smart home control: invalid actions and incorrect interpretation of user requests. These failures motivate a design that separates semantic interpretation of the user requests from runtime enforcement of action constraints and supports abstention when a request cannot be resolved from the available context.

\noindent{\textbf{Example Setup.}} Consider a smart home with four devices: a) {light bulb} in the {living room}, b) {thermostat} in the {living room}, c) {light bulb} on the {balcony}, and d) a {lock} on the {front door}. Each light bulb can be turned on or off, and configured to a brightness level 0--100. The thermostat can be set to 16--30$^\circ$C, and the front door can be locked or unlocked. We use Qwen2.5 1.5B, 3B and 7B to translate user requests to device actions within this home -- Table~\ref{tab:model-outputs} shows the requests and outputs that we refer to in this section.

\noindent{\textbf{Failure Type 1: Invalid Action Failures.}} Consider request \textbf{\circled{3}} -- the 7B model generates \code{bedroom.AC.on()}, which is a reasonable output if the smart home contained the device. Similarly, request \textbf{\circled{5}} specifies an invalid brightness level -- the 1.5B model generates \code{balcony.light.set_brightness(150)}, placing the invalid value into the actuation command. The 3B model makes an additional mistake by introducing a syntax error (: instead of . after light). 

These failures show that SLMs do not consistently enforce device capabilities, argument bounds, or command syntax. Such constraints should instead be enforced by the runtime.

\noindent{\textbf{Failure Type 2: Interpretation Failures.}} A valid output can still misinterpret a user's request. The request \textbf{\circled{4}} is ambiguous because the example home contains two light bulbs. All SLMs turned off the living room light, despite insufficient information to determine that this is the user's intended target. Similarly, request \textbf{\circled{2}} requires interpreting how much to increase the temperature based on the current state and user preferences. 

Unlike invalid actuations, these failures cannot be prevented by enforcing device constraints alone. The model must resolve the user's intent from the available context or abstain when the request remains ambiguous.

\noindent{\textbf{Failure Amplifiers.}}
These failure modes are further exacerbated by i) compound requests, which contain more than one command in the same user request (for e.g., \textbf{\circled{6}}), and ii) convoluted natural-language expressions like slang expressions e.g. \texttt{``the heat has me cooked''} which means the user wants to lower the temperature, not actually cook. 

\noindent{\textbf{Implication.}} These failures are particularly consequential in smart homes, where incorrect commands can affect physical devices. For request \textbf{\circled{7}}, all SLMs generate invalid commands that could leave the front door unlocked (possibly even at night). While not unique to SLMs, such failures highlight the limitations of relying on unconstrained command generation for reliable smart home control. 
\section{\name{}'s Design}
\label{sec:design}

\name{} translates natural-language requests into executable smart home API calls through a hierarchy of typed decisions grounded in the home's device registry. This section describes its design, summarized in Algorithm~\ref{alg:mainalgo}. More details about the typed decisions used in \name{} are in Appendix \ref{app:heartprompts}.

\subsection{Smart Home Control Problem}
\name{} formulates smart home control as a selection problem where a smart home system selects actions to satisfy a user's request. The space of possible actions is large -- a naive selection considers $RDO$ candidates for a home with $R$ rooms, $D$ devices per room, and $O$ operations per device, and may produce invalid device calls. This action space also varies across homes and changes as devices are added or removed. In contrast, \name{} considers only $R+D+O$ candidates while ensuring that the selected device and operation are valid.

We define a smart home $H$ by its device registry $R_H$ and device state $S_H$. $R_H$ defines $H$'s closed set of available $operations$ and their parameter signatures. $S_H$ records each device's current status and attributes. Fig. \ref{fig:home-example} shows an example $R_H$ and accompanying $S_H$. Let $A_H$ denote the set of valid calls to these operations with arguments in their declared domains.

Given a user request $u$, that may contain several requests $s$ (for e.g., ``dim the balcony light to 50\% and make it 2 degrees cooler''), smart home control maps $u$, $R_H$, and $S_H$  to an ordered list of actions in $A_H$ or refusals. The refusal is required when the home cannot do what was asked, when the request does not say enough to act, or when the request asks for a value the device cannot take. The correct output is a list with one item per single request.

\begin{figure}[h]
\centering
\fbox{%
\begin{minipage}{0.93\columnwidth}
\footnotesize
\textbf{Registry $R_H$: available operations}
\begin{flushleft}
\ttfamily
living\_room.light.\\
\hspace*{1em}turn\_on()\\[2pt]
living\_room.light.\\
\hspace*{1em}set\_brightness(brightness:int)\\[2pt]
living\_room.thermostat.\\
\hspace*{1em}set\_temperature(temp:int)
\end{flushleft}

\medskip
\textbf{State $S_H$: current values and domains}
\begin{flushleft}
\ttfamily
living\_room.light:\\
\hspace*{1em}status = on\\
\hspace*{1em}brightness = 60\\
\hspace*{1em}brightness range = [0,100]\\[4pt]
living\_room.thermostat:\\
\hspace*{1em}temperature = 21\\
\hspace*{1em}temperature range = [16,30]
\end{flushleft}
\end{minipage}%
}
\caption{Example registry $R_H$ and device state $S_H$ for a home $H$ with a living room light and thermostat.}
\label{fig:home-example}
\end{figure}

\subsection{System Design}

\begin{algorithm}[h]
\caption{\name{} resolves each request to an API call or a refusal using hierarchical typed decisions.}
\label{alg:homing}
\begin{algorithmic}[1]
\REQUIRE Request $u$, home registry $R_H$, device state $S_H$, Jev selector $D$, confidence threshold $\tau$
\ENSURE An ordered list with one valid API call or \code{error_input} per sub-command $s$
\STATE $T \gets \textsc{Tree}(R_H)$ \COMMENT{room $\to$ device $\to$ operation}
\FORALL{sub-command $s \in \textsc{Split}(u)$}
  \STATE $n \gets \textsc{Root}(T)$;\; $\textit{path} \gets []$;\; $c \gets 1$
  \WHILE{$n$ is not an operation leaf}
    \STATE $C \gets \textsc{Children}(n)$ \COMMENT{available rooms, devices, or operations}
    \STATE $O \gets \{\textsc{Name}(k) : k \in C\} \cup \{\textsc{none}\}$
    \STATE $Q \gets \textsc{Question}(n)$ \COMMENT{which room, device, or operation?}
    \STATE $p \gets D(\textsc{Context}(s, \textit{path}), Q, O)$ 
    \STATE $k^* \gets \arg\max_{k \in C} p(\textsc{Name}(k))$ \COMMENT{best non-\textsc{none} choice}
    \STATE $c \gets c \cdot p(\textsc{Name}(k^*))$ \COMMENT{do not renormalise after excluding \textsc{none}}
    \STATE Append $k^*$ to $\textit{path}$
    \STATE $n \gets k^*$ \COMMENT{descend to the selected child}
  \ENDWHILE
  \IF{$c < \tau$}
    \STATE Emit \code{error_input} \COMMENT{low confidence in the complete action path}
  \ELSE
    \STATE Emit $\textsc{SelectArgument}(s, n, S_H, D)$
  \ENDIF
\ENDFOR
\end{algorithmic}

\label{alg:mainalgo}
\end{algorithm}

\noindent{\textbf{Hierarchical Action Resolution.}} \name{} needs to convert each natural language request into a structured query. To do so, \name{} organises $R_H$ as a tree. Each registry entry in $R_H$ has a dotted name \code{room.device.operation(arguments)} and a signature that lists its parameters with their types, for e.g., \code{hall.heat.set_temp(temp:int)}.  In the tree, each room branches into its devices, and each device branches into its supported operations. Selecting a room narrows the candidate devices, and selecting a device narrows the available operations. A path from the root to a leaf therefore identifies an executable operation in the home.

The three questions that \name{} asks to resolve all user requests are: (i) \emph{Which room does this command refer to?} (ii) \emph{Which device in that room does this command control?} and (iii) \emph{Which operation on that device does this command ask for?} Consider the request ``dim the living room light to 40\%''. At the root, \name{} asks which room the request refers to, offering the rooms in the registry as choices. Selecting \code{living_room} exposes only the devices in that room.  Whole-home devices, such as a vacuum robot, are offered as choices within every room. The next question identifies \code{light}, whose available operations then become the choices for the final question. Selecting \code{set_brightness} completes the path.

As its selector for each level, \name{} deploys a singular instance of JevK5 \cite{openjev, almeida2026systemone}, an open-sourced Decision Model finetuned to hierarchical smart home questions . The selector grounds each request in the registry through \emph{typed decisions}: each decision selects from a predefined set of options corresponding to rooms, devices, or operations. At each of the three levels, the selector sees the original request, the selections made so far, and the current node's children -- it never needs to consider all $R_H$ actions in a single decision. The selector returns a probability distribution over the options, which \name{} uses both to select the next branch and to accumulate confidence in the resulting action path. This path selection occurs independently for each sub-command $s$ in the user request $u$.

\noindent{\textbf{Deciding When to Accept or Refuse.}} Restricting choices to the registry prevents invalid operations (Failure Type 1, \S\ref{sec:slmfailure}) but does not guarantee that the selected action matches the user's intent (Failure Type 2). Unlike LLMs that may persist even when a task is impossible, \name{} can recognize when to give up. To enable this, each hierarchical typed decision includes a \emph{none} option, allowing the selector to assign probability to the possibility that no available choice matches.

\name{} continues through the highest-probability real option, retaining its probability without renormalising away \emph{none}, even when \emph{none} scores higher. It accumulates confidence along the path: $c(s) =p(r \mid s)p(d \mid s,r)p(o \mid s,r,d)$,
where $r$, $d$, and $o$ are the selected room, device, and operation, respectively, and $p$ is the probability assigned by the selector given the request and preceding selections.

After all three hierarchical decisions and reaching an operation, \name{} refuses if $c(s) < \tau$; otherwise, it proceeds to resolve the operation's arguments. This defers the refusal decision until the complete path has been evaluated, ensuring that one uncertain level doesn't decide if the user request can be resolved or not.

\noindent{\textbf{Argument Resolution.}} Selecting a device and operation is not sufficient to execute commands that require parameters, such as setting a temperature or brightness level. \name{} must also determine the intended value while ensuring it satisfies the device's constraints. To do so, \name{} uses the operation's parameter name and type from the registry and its allowed range from the device state.

If the final action takes no parameter (for e.g., \code{living_room.light.turn\_on()}), the request resolution is complete. However, when $u$ states a numeric target, \name{} parses it directly. Relative changes are computed directly using the current value in $S_H$ -- ``2 degrees cooler'' becomes $21-2=19$ when the current temperature is $21^\circ$C. When $u$ leaves the value unspecified, as in ``make it a bit warmer'', \name{} uses a coarse-to-fine selection procedure. For an integer domain $[\ell,h]$, it first offers the selector a coarse set of evenly spaced candidate values. 
After each selection, \name{} narrows the interval around the selected value to resolve the argument. All candidates are restricted to the declared domain and an out-of-range value produces a refusal. Finally, for enumerated parameters such as operating modes, the selector instead chooses directly among the declared options.

\subsection{Memory Module}

Natural-language requests often rely on user-specific context that cannot be inferred from the request or the home's current state. For example, ``turn off the front light'' requires knowing which device the user calls the ``front light,'' while ``set the lights to my usual brightness'' requires knowing the user's preferred setting. To capture these two types of information, \name{} maintains a structured memory of \emph{preferences} and \emph{references}, stored separately for each user in a household.

\emph{Preferences} associate device operations with preferred values, optionally conditioned on a situation, e.g., setting the living-room light to 40\% when reading or the dehumidifier to 100\% when it rains. \emph{References} map user-defined names like ``the front light'' to devices and track recently operated devices, enabling follow-up requests such as ``turn it off.'' 

This design integrates naturally with hierarchical action resolution: references identify the intended device before operation selection, while preferences supply missing values afterward. Memories are stored locally in a SQLite database.

\noindent{\textbf{Routing.}}
For each sub-command $s$, \name{} first uses a typed decision to determine whether the user is requesting an immediate action, providing information to remember, or revising stored information or a previous action. The decision returns a probability distribution over three choices: \code{now}, \code{remember}, and \code{revise}. Since both \code{remember} and \code{revise} require memory processing, \name{} routes the sub-command to the write path when $1 - P(\texttt{now} \mid s) \geq \tau_m$.

\noindent{\textbf{Writing Memory.}}
For sub-commands routed to memory, a second typed decision identifies one of four operations: \code{preference}, \code{alias}, \code{forget}, or \code{correction}. Note that \code{alias} and \code{correction} both create references, while \code{forget} removes references or
preferences. A preference stores a preferred value for a device operation (``I prefer the bedroom light at 40\%''); a reference names a device (``call the hallway light the front light''); a forget request removes stored aliases or preferences; and a correction, when the previous action targeted the wrong device, re-runs that command on the intended device and remembers the user's phrase for it.

Each memory operation uses \name{}'s hierarchical action resolution to identify the relevant device. Typed decisions then select the information to store from candidate phrases drawn directly from the user's message. For a \emph{preference}, \name{} takes the operation from this resolution, parses the preferred value from the message, e.g., brightness at 40\%, and identifies any situational context by selecting relevant phrases from the message. For an \emph{alias}, it selects the user-defined device name. For a \emph{correction}, it selects the device reference in the earlier request, associates it with the corrected device, and re-executes the request. A \emph{forget} operation removes the references the message mentions, or else the preferences of the identified device, from active memory. If the required information cannot be identified, \name{} requests clarification without updating memory.

\noindent{\textbf{Reading Memory.}} When the command is not a memory write, \name{} reads memory before and after the hierarchical action resolution. Consider a user who earlier said ``call the hallway light the front light'' and ``I like the bedroom light at 40\% when I'm reading''.

\underline{Before resolution, memory can fix the device.} For ``turn on the front light'', the saved name ``front light'' appears in the message, so the device is fixed to \code{hallway.light} and only the operation is chosen (\code{turn_on}): one typed decision instead of three. For a follow-up ``now turn it off'', no saved name appears and no device is named, but the message refers back (``it''), so one typed decision chooses among the devices acted on in the last three turns, or ``none of these''. A device named in the message is never overridden: with no matching name and no reference back, resolution runs over the whole home exactly as without memory.

\underline{After resolution, memory can supply the value.} For ``set the bedroom light to my usual, I'm reading'', the resolution selects \code{bedroom.light.set_brightness} but the message states no value. \name{} looks up the saved preferences for that device and operation, finds the one whose situation (``reading'') appears in the message, and fills in 40. A value stated in the message (``to 60\%'', ``a bit brighter'') always takes precedence over a preference.

Finally, the action is checked against the home's registry, the user's permissions and the device's allowed values, and is refused if any check fails; if saved names for several devices match, \name{} asks the user which one is meant.

\section{Training and Calibrating \name{}}
\label{sec:implementation}

\name{}'s design (\S\ref{sec:design}) relies on a selector that answers its typed questions and on three values that control its behavior: a softmax temperature $T$ for the selector's probabilities, the refusal threshold $\tau$, and the memory gate threshold $\tau_m$. This section describes the data used to obtain them. 

\subsection{Adapting HomeBench}
\label{sec:adaptedhb}

To train (and later, evaluate) \name{} to translate natural-language requests into valid device actions across diverse smart homes, we adapt HomeBench~\cite{homebench}. HomeBench pairs user requests with executable device calls across 100 simulated homes, each exposing approximately 140 actions through its device registry. Each request $u$ contains up to ten sub-commands $s$, each mapped to a device call or \code{error_input} if the home cannot execute it.

HomeBench's requests are almost always explicit: they name the room, the device, the operation and the target value, as in \emph{``Lower the temperature of the air conditioner in the master bedroom to 20 degrees.''} Such requests mainly test whether a system can map a fully specified instruction to an API call. In everyday requests, users may leave these details implicit: \emph{``make the master bedroom cooler''}. We therefore use GPT-5 to rewrite HomeBench's requests into four styles. 
\begin{itemize}[
    leftmargin=*,
    topsep=0pt,
    itemsep=0pt,
    parsep=0pt,
    partopsep=0pt
]
  \item \textbf{Original}: the original HomeBench request (22.5\%) 
  \item \textbf{Explicit}: colloquial but fully specified (22.5\%)
\item \textbf{Implied Device}: device inferred from context (45\%)
\item \textbf{Goal Only}: desired outcome, not action (10\%)
\end{itemize}

The resulting corpus contains natural-language requests paired with action or refusal labels. Note that we do not make any changes to the home configurations or the ground truth provided. We drop rows where the ground truth was incorrect (undefined action or out-of-range values). The prompts used are provided in Appendix \ref{app:prompt}.

\subsection{Selector Finetuning}

\name{} uses a finetuned selector for all of its typed decisions: the three levels of action resolution, argument selection, and memory operations. We finetune JevK5 \cite{openjev} once, on action-resolution and argument decisions only, and reuse it for memory decisions without further training. Because the registry and the memory supply the options of every question at runtime, the finetuned selector needs no further training for a new home.

\noindent{\textbf{Constructing Decision Examples.}}
Since \name{}'s selector must answer hierarchical questions rather than generate complete device calls, we finetune it on individual typed decisions derived from the adapted HomeBench. We turn each labeled request into the individual choices the selector makes during inference. For example, an action labeled \code{living_room.light.set_brightness(40)} provides at least three examples: selecting the living room, selecting its light, and selecting the brightness operation. Each example pairs the request and the correct preceding selections with a question, the available options, and the correct answer. The user request stays the same.

For refusal requests, we label the first unmatched decision as \emph{none}; for example, ``turn on the living room fan'' receives \code{living_room} for the room decision and \emph{none} for the device decision if no fan exists. Actions with arguments provide additional examples. For an enumerated argument, the correct answer is the labeled option. For a numeric argument, the coarse selection targets the candidate closest to the labeled value, splitting the target probability equally if two candidates tie. The fine selection then targets the exact labeled integer among the nearby values.

\noindent{\textbf{Dataset Composition.}}
The decision examples yield about 250{,}000 unique examples. We sample 100{,}000 decisions from this pool, allocating 25{,}000 each to room, device, operation, and argument selection. At each hierarchy level, refusal examples occupy up to half the budget, with positive examples filling the remainder. Within the argument category, enumerated, coarse, and refinement examples are sampled in proportion to their availability. 

\noindent{\textbf{Training Configuration.}} We initialise the selector from JevK5, a Qwen3.5-4B checkpoint trained for typed decisions. We finetune it for one epoch using rank-16 LoRA, a learning rate of $3\times10^{-5}$, 20 warm-up steps, and an 8{,}192-token batch budget per GPU across four GPUs.

\subsection{Calibrating the Selector}

\name{} uses the selector's probabilities to determine when to abstain, making calibration important for reliable refusal decisions. We apply temperature scaling, which adjusts confidence without changing the selected option. We fit $T=1.0$ by minimizing negative log-likelihood on half of approximately 7{,}000 held-out decisions spanning room, device, operation, refusal, and value questions, and evaluate it on the remaining half. 

\noindent{\textbf{Setting the Refusal Threshold.}} \name{} refuses a sub-command when its path confidence $c(s)$ is below $\tau$. We use validation requests to set $\tau=0.27$.

\noindent{\textbf{Setting the Memory Gate.}} A typed decision sends memory statements to the write path using $\tau_m$. We use validation requests to set $\tau_m=0.74$ .

\section{Evaluation Datasets}
\label{sec:datasets}

We build 3 smart home datasets for evaluating \name{}.

\subsection{Trek-Base: Test Set}
We evaluate \name{} on the test split of the adapted HomeBench (\S\ref{sec:adaptedhb}). The test set contains 15{,}476 requests over 100 homes.
The split mixes single requests that the home supports (5{,}732), single requests for a device the home does not have (4{,}063) or an attribute the device does not have (2{,}523), and 3{,}158 utterances that combine 2 to 10 requests. We drop requests with inconsistent ground truths, leaving 14{,}166 valid requests. 
Because HomeBench's train and test splits share homes and sentence templates, this dataset measures performance on familiar homes and familiar phrasing.

\subsection{Trek-Robust: Robustness Evaluation}
Trek-Robust tests whether a system still works when the home or the wording changes. We introduce 9 categories that change either the user request or the home: \emph{counterfactual} -- the same user request in two slightly different homes so that the correct answers differ, \emph{unseen capability} -- the device lacks the requested operation, has a new operation signature, or does not declare what it can do, \emph{missing information} -- the needed capability is hidden or unknown, so the correct answer is a refusal, \emph{novel identity} -- new names, aliases or descriptions, or an added device of the same type, \emph{new home} -- 20 new homes with new rooms and device types, \emph{new vocabulary} -- another ecosystem's command grammar like Home Assistant-style service calls, vendor-prefixed methods, REST requests, or HomeBench's format with synonym operation names, e.g., \texttt{set\_level} for \texttt{set\_brightness}, \emph{rendering} -- the same home in other formats like JSON, YAML, shuffled rooms, etc., \emph{registry scale} -- homes enlarged to about 300, 600 or 1{,}000 operations while keeping the existing request, and \emph{state dependence} -- a change by a stated amount that stays in range in one state and leaves it in another. 

All categories start from an Trek-Base request. 6 categories were generated with GPT-5 (prompts in Appendix \ref{app:Trek-Robustprompt}); the remaining 3 were built deterministically (rendering, registry\_scale, and state\_dependent). 
Trek-Robust has about 2{,}000 valid rows: 200 per category, and about 400 for counterfactual (roughly 200 complete pairs).

\subsection{Trek-Memory: Conversational Memory Evaluation}
Trek-Base and Trek-Robust score each request on its own. To test memory across turns, we build the Trek-Memory dataset. It groups the Trek-Base requests into 100 multi-turn conversations, one per home, each with a single user, and adds requests that need memory. Each home's requests are shuffled once. After each turn, the ground truth actions are applied to the simulated home, so later turns see the updated state. Relative requests (e.g., ``turn it up by 2'') are answered from this updated state, including changes made by the added turns. Requests whose correct answer can no longer be determined are dropped.

We add about 30 memory turns to each home, 2{,}895 in total. They are built from templates and each home's devices:
\begin{itemize}[
    leftmargin=*,
    topsep=0pt,
    itemsep=0pt,
    parsep=0pt,
    partopsep=0pt
]
    \item \textbf{Preferences.}
    We assign two preferences to each of the 100 homes. The user states the preference (``My usual brightness for the balcony light is 30\%'') and later asks for it (``Set the balcony light to my usual brightness''). The first preference tests situation-specific values, explicit-value overrides, and subsequent recall. The second tests using, updating, and forgetting a preference, with a final request requiring refusal.

    \item \textbf{Alias.}
    Two devices per home receive nicknames (e.g., ``Bubbles'' and ``the tall fan''), each used twice. The user then forgets one nickname and reuses it, or uses an undefined nickname, both requiring refusal.

    \item \textbf{Corrections.}
    A request uses a phrase that matches no device. The user then names the intended device (``No, I meant the store room light''), and a later request reuses the phrase for that device.

    \item \textbf{Follow-ups.}
    Twice per home, a request that names a device is followed right away by one that refers to it as ``it'' or ``its''.
\end{itemize}

Each group of memory turns starts in the first half of the conversation, with a random number of original requests between its turns. We then reword the added turns with GPT-5 (prompt in Appendix \ref{app:Trek-Memoryprompt}), one call per home, so they do not all share the template wording. Rewording changes only the text; ground truth answers and turn order stay the same. Memory has 14{,}906 turns turns across 100 conversations, with a median of 182 turns per conversation. 

\section{Evaluation}

We evaluate \name{} on the following research questions:
\begin{enumerate}[
    leftmargin=*,
    topsep=0pt,
    itemsep=0pt,
    parsep=0pt,
    partopsep=0pt
]
    \item How accurate is \name{}, and how robust is it to variations in homes and request phrasing (\S\ref{sec:eval:main})?
\item How efficient is \name{} in terms of latency (\S\ref{sec:eval:latency})?
\item How well does \name{} generalise to unseen homes (\S\ref{sec:eval:gen})?
\item Can \name{} leverage conversational memory and state across multiple turns (\S\ref{sec:eval:memory})?
    
\end{enumerate}

\noindent{\textbf{Baselines.}} We compare \name{} against Qwen3.5-4B (i.e. JevK5's decision engine) with prompting and finetuning on approximately 22{,}000 adapted HomeBench requests, VCU-LLM~\cite{vcullm}, Sasha~\cite{sasha}, AdaHome~\cite{adahome}, and Domux~\cite{domux}. VCU-LLM uses a finetuned Llama-3.2-1B with a 119-request retrieval bank, a hand-written verb dictionary, and a validation-tuned retrieval threshold. Sasha and AdaHome use Qwen3.5-4B. We use Domux's released weights and reproduce VCU-LLM, Sasha, and AdaHome from their papers, reusing available prompts and mapping Domux's predicted slots to the home's registry. Qwen3.5-4B uses the prompt from HomeBench \cite{homebench}. The prompts can be found in Appendix \ref{app:baselineprompt}.

\noindent{\textbf{Metrics.} We report strict exact match accuracy: a request is correct only if the multiset of actions a system outputs equals the ground truth, where each item is a complete device call or a refusal (\code{error\_input}); a request with several sub-commands is correct only if every one is. This strict criterion reflects the need for reliable smart home control, where even a single incorrect sub-command can leave the user's request partially unfulfilled or cause unintended changes to the home.

\subsection{Accuracy and Robustness}
\label{sec:eval:main}

\begin{table}[h]
\centering
\small
\setlength{\tabcolsep}{3pt}
\begin{tabular*}{\columnwidth}{@{\extracolsep{\fill}}lccc@{}}
\toprule
 & Trek-Base & Trek-Robust \\
\midrule
\name{} & 0.862 & \textbf{0.853} \\
\midrule
Qwen3.5-4B, prompted & 0.495 & 0.451 \\
Qwen3.5-4B, finetuned & \textbf{0.897} & 0.698 \\
VCU-LLM & 0.849 & 0.505 \\
Domux & 0.517 & 0.467 \\
Sasha & 0.463 & 0.463 \\
AdaHome & 0.306 & 0.329 \\
\bottomrule
\end{tabular*}
\caption{Exact match accuracy on Trek-Base and Trek-Robust.}
\label{tab:main}
\end{table}

\emph{\name{} trades some accuracy on familiar requests for greater robustness to variations in homes.} Table \ref{tab:main} reports the accuracy of \name{} and the baselines on Trek-Base and Trek-Robust. On Trek-Base, finetuned Qwen3.5-4B achieves the highest accuracy at 89.7\%, followed by \name{} at 86.2\%. However, in Trek-Robust, the finetuned generator drops by 19.9 percentage points to 69.8\%, while \name{} drops by only 0.9 points, making 2.1$\times$ fewer errors. Although both \name and Qwen3.5-4B are fine-tuned, they learn different capabilities. Qwen3.5-4B learns to map requests directly to device calls, performing well on familiar homes and phrasing. In contrast, \name selects rooms, devices, and operations from the current home's registry, including refusal when appropriate. This makes \name more robust to changes in homes and requests, as tested by Trek-Robust.

\begin{table}[h]
\centering
\small
\setlength{\tabcolsep}{3pt}
\resizebox{\columnwidth}{!}{%
\begin{tabular}{@{}lccccc@{}}
\toprule

& Original
& Explicit
& \shortstack{Implied}
& \shortstack{Goal}
& All \\
\midrule
\name{} & 0.994 & 0.885 & 0.817 & 0.703 & 0.862 \\
\midrule
Qwen3.5-4B, prompted & 0.788 & 0.424 & 0.442 & 0.196 & 0.495 \\
Qwen3.5-4B, finetuned & \textbf{0.998} & \textbf{0.913} & \textbf{0.865} & 0.771 & \textbf{0.897} \\
VCU-LLM & 0.959 & 0.843 & 0.801 & \textbf{0.795} & 0.849 \\
Domux & 0.675 & 0.417 & 0.483 & 0.477 & 0.517 \\
Sasha & 0.620 & 0.342 & 0.446 & 0.406 & 0.463 \\
AdaHome & 0.469 & 0.234 & 0.276 & 0.203 & 0.306 \\
\bottomrule
\end{tabular}%
}
\caption{Exact match accuracy on Trek-Base by wording style.}
\label{tab:styles}
\end{table}

\noindent{\textbf{Accuracy by Request Wording.}} \emph{\name{} reliably handles explicit commands but loses accuracy as phrasing becomes more colloquial or intent less specified, like most baselines.} Table~\ref{tab:styles} shows \name{} dropping from 99.4\% on original requests to 88.5\% on colloquial, 81.7\% on implied-device, and 70.3\% on goal-only requests. Finetuned Qwen3.5-4B outperforms \name{} across all styles, while VCU-LLM achieves the highest goal-only accuracy (79.5\%).

\begin{table*}[t]
\centering
\small
\setlength{\tabcolsep}{4pt}
\begin{tabular*}{\textwidth}{@{\extracolsep{\fill}}lcccccccccc@{}}
\toprule
 & CF & UC & MI & NI & Rnd & NH & NV & Scale & SD & All \\
\midrule
\name{} & \textbf{0.851} & \textbf{0.670} & \textbf{0.885} & \textbf{0.920} & 0.855 & \textbf{0.685} & \textbf{0.915} & 0.905 & \textbf{0.995} & \textbf{0.853} \\
\midrule
Qwen3.5-4B, prompted & 0.551 & 0.375 & 0.125 & 0.570 & 0.450 & 0.555 & 0.305 & 0.550 & 0.475 & 0.451 \\
Qwen3.5-4B, finetuned & 0.799 & 0.620 & 0.245 & 0.915 & \textbf{0.910} & 0.670 & 0.600 & \textbf{0.950} & 0.470 & 0.698 \\
VCU-LLM & 0.630 & 0.250 & 0.310 & 0.405 & 0.860 & 0.115 & 0.875 & 0.815 & 0.155 & 0.505 \\
Domux & 0.479 & 0.380 & 0.690 & 0.510 & 0.580 & 0.265 & 0.460 & 0.670 & 0.155 & 0.467 \\
Sasha & 0.434 & 0.325 & 0.630 & 0.420 & 0.535 & 0.385 & 0.485 & 0.535 & 0.450 & 0.463 \\
AdaHome & 0.365 & 0.315 & 0.020 & 0.425 & 0.245 & 0.505 & 0.310 & 0.365 & 0.375 & 0.329 \\
\bottomrule
\end{tabular*}
\caption{Exact match accuracy on Trek-Robust by categories. CF: counterfactual;
UC: unseen capability; MI: missing information;
NI: novel identity; Rnd: rendering; NH: new home;
NV: new vocabulary; Scale: registry scale;
SD: state dependence; All: accuracy over all requests.}
\label{tab:Test-Hard}
\end{table*}

\noindent{\textbf{Robustness Across Trek-Robust Categories.}} \emph{\name{} outperforms the baselines on most categories in Trek-Robust, especially when actions depend on runtime device state, capability information, and API vocabulary, while also achieving the highest accuracy on new homes and unseen capabilities.} Table~\ref{tab:Test-Hard} presents the per-category results on Trek-Robust. For state-dependent requests (SD), \name{} achieves 0.995 accuracy by computing the result directly from device state, whereas the strongest baseline reaches only 0.475. When capability information is missing or unknown (MI), the correct response is a refusal. \name{} correctly refuses 0.885 of these requests, substantially outperforming Sasha (0.630) and Domux (0.690). When the API vocabulary changes (NV), \name{} and VCU-LLM maintain high accuracy (0.915 and 0.875) by selecting operations from the runtime registry, while the finetuned generator, trained to produce HomeBench-style calls, achieves only 0.600. \name{} also achieves the highest accuracy on new homes (NH, 0.685) and unseen capabilities (UC, 0.670), although these remain its lowest-scoring categories, indicating that unfamiliar configurations and capabilities are more difficult to handle.

The finetuned generator performs best on registry rendering (Rnd, 0.910) and scale (Scale, 0.950), where the underlying requests and operations remain largely familiar. Counterfactual requests (CF) further highlight this distinction: the same instruction is evaluated under different home configurations that require different outputs. \name{} achieves the highest accuracy (0.851), outperforming the finetuned Qwen3.5-4B (0.799). This suggests that the finetuned generator relies more heavily on request-to-action patterns learned during training, whereas \name{} better adapts its decisions to changes in the runtime environment.

\subsection{Latency}
\label{sec:eval:latency}

\begin{figure*}[t]
  \centering
  \begin{subfigure}{0.32\textwidth}
    \centering
    \includegraphics[width=\linewidth]{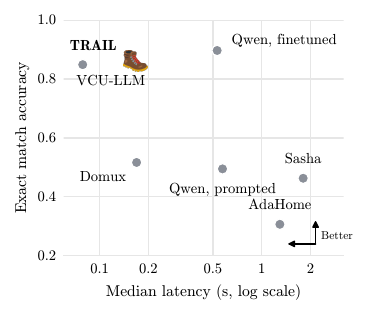}
    \caption{Trek-Base}
    \label{fig:accuracy-latency-Trek-Base}
  \end{subfigure}
  \hfill
  \begin{subfigure}{0.32\textwidth}
    \centering
    \includegraphics[width=\linewidth]{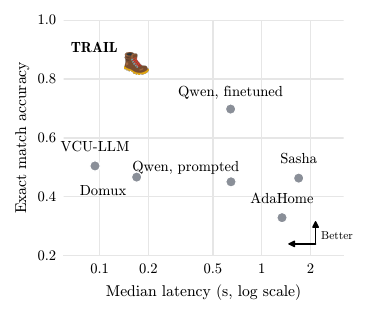}
    \caption{Trek-Robust}
    \label{fig:accuracy-latency-Trek-Basehard}
  \end{subfigure}
  \hfill
  \begin{subfigure}{0.32\textwidth}
    \centering
    \includegraphics[width=\linewidth]{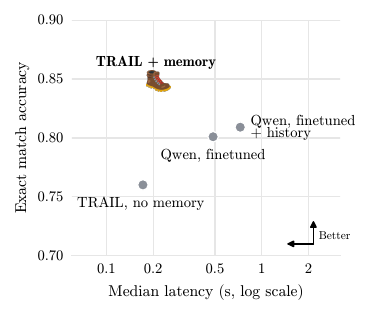}
    \caption{Trek-Memory}
    \label{fig:accuracy-latency-Trek-Memory}
  \end{subfigure}
  \caption{Exact match accuracy versus median latency. Up and to the left is
  better.}
  \label{fig:accuracy-latency}
\end{figure*}

\emph{\name{} achieves low inference latency despite requiring multiple typed decisions per request, and maintains this advantage when conversational memory is enabled.} Fig. \ref{fig:accuracy-latency} plots the end-to-end latency of \name{} and its baselines on Trek-Base, Trek-Robust and Trek-Memory (the exact results are in Tables \ref{tab:latency_Trek-Base}, \ref{tab:latency_Trek-Robust} and \ref{tab:memory-latency} in the Appendix \ref{app:latency}). For each dataset, every system answers the same  requests (500 for Trek-Base and Trek-Robust, 575 i.e. 3 conversations for Trek-Memory), one at a time as a smart home assistant would serve them. All systems run on a single NVIDIA RTX A4500 GPU. \name{} runs with Hugging Face Transformers; the LLM baselines run with vLLM, which is heavily optimized.

\name{} achieves a median latency of 0.166 s on Trek-Base and 0.169 s on Trek-Robust, making it 3.2$\times$ and 3.8$\times$ faster than its competitive baseline, the finetuned Qwen3.5-4B, respectively. This is despite \name{} requiring an average of 5.4 and 4.5 typed decisions per request, compared with a single generation call for Qwen3.5-4B. VCU-LLM achieves the lowest median latency (0.079\,s on Trek-Base and 0.094\,s on Trek-Robust), but its accuracy drops substantially on Trek-Robust (0.505 versus 0.853 for \name{}). Domux matches \name{}'s median but is far less accurate.

Enabling memory increases \name{}'s median latency from 0.172 s to 0.215 s (25\%) and the average number of typed decisions from 5.7 to 7.5 per turn. In comparison, providing the full conversation history to finetuned Qwen3.5-4B increases its median latency from 0.486 s to 0.727 s. With memory enabled, \name{} remains 3.4$\times$ faster at the median and 1.6$\times$ faster at P90 than Qwen3.5-4B with history.

\subsection{Generalization}
\label{sec:eval:gen}

\begin{table}[H]
\centering
\small
\resizebox{\columnwidth}{!}{%
\begin{tabular}{@{}lccc@{}}
\toprule
\# Homes Training & Trek-Base, 20 left-out & Trek-Robust, new home \\
\midrule
100 (\name{}) & 0.859 & 0.685 \\
80 & 0.860 & 0.695 \\
\bottomrule
\end{tabular}%
}
\caption{\name{} trained on the requests of all 100 homes, or of 80 homes with the other 20 left out (randomly chosen).}
\label{tab:heldout}
\end{table}

\emph{\name{} grounds its decisions in the home's runtime device registry, allowing it to operate on unseen home configurations without home-specific training.} To evaluate this, we retrain \name{} on 80 homes, excluding all requests from the remaining 20. As Table~\ref{tab:heldout} shows, the model trained on 80 homes achieves nearly identical accuracy on the 20 held-out homes as the model trained on all 100 homes (0.860 versus 0.859). It also performs comparably on Trek-Robust's new-home category (0.695 versus 0.685), which contains 20 generated homes unseen by both models. These results suggest that \name{} learns to select actions from the runtime registry rather than relying on memorized home-specific configurations, enabling generalization to unseen homes without retraining.

\subsection{Memory}
\label{sec:eval:memory}

\begin{table}[H]
\centering
\small
\setlength{\tabcolsep}{4pt}
\resizebox{\columnwidth}{!}{%
\begin{tabular}{@{}lccc@{}}
\toprule
 & All & Original & Memory \\
\midrule
\name{}, no memory & 0.760 & 0.872 & 0.277 \\
\name{} + memory & 0.848 & 0.870 & 0.756 \\
\midrule
Qwen3.5-4B, finetuned & 0.801 & 0.905 & 0.350 \\
Qwen3.5-4B, finetuned + history & 0.809 & 0.898 & 0.425 \\
\bottomrule
\end{tabular}%
}
\caption{Exact match accuracy on Trek-Memory.
\emph{Original}: Trek-Base requests with state updates;
\emph{Memory}: Added memory turns.
\emph{History}: The full conversation history is included in the prompt.}
\label{tab:memory}
\end{table}

\emph{Structured memory substantially improves \name{}'s ability to resolve conversational requests without degrading performance on ordinary requests, outperforming finetuned Qwen3.5-4B even when provided with the full conversation history.} Table~\ref{tab:memory} shows the accuracy on Trek-Memory. Adding memory increases accuracy on memory-dependent turns from 0.277 to 0.756, while accuracy on original Trek-Base requests remains nearly unchanged (0.872 versus 0.870). Providing the full conversation history to finetuned Qwen3.5-4B improves its memory accuracy from 0.350 to 0.425, but it still falls substantially behind \name{} (0.756). Although the finetuned generator performs better on original requests (0.898 versus 0.870), \name{} achieves higher overall accuracy (0.848 versus 0.809).

\section{Related Work}

\subsection{Natural-Language Smart Home Control}

Recent works improve natural-language driven smart home control using model adaptation, retrieval and planning. VCU-LLM \cite{vcullm} combines a finetuned Llama3.2-1B SLM with knowledge retrieval and KV cache optimization. MiCU \cite{micu} uses curriculum learning, chain of thought reasoning, and reinforcement learning, while Domux \cite{domux} finetunes a Gemma-4-E2B SLM to convert user commands into structured, pipe-delimited slots. AdaHome \cite{adahome}  routes commands to either a direct or a Chain-of-Draft reasoning planner based on the request complexity. 

Other smart home control systems emphasize goal interpretation and tool orchestration. SAGE \cite{sage} grounds user requests in smart home actions through LLM-driven tool use such as: device disambiguation, API inspection, preference retrieval, and device command execution. Similarly, Sasha \cite{sasha} uses an iterative reasoning process to produce action plans to respond to vague user commands.

These systems establish the value of domain specific training and context for smart home interaction. \name{} focuses on a different inference mechanism: composing typed candidate decisions over a runtime-constructed device hierarchy rather than generating complete commands or action plans.

\subsection{Grounded Action Selection and Hierarchical Predictions}

Grounding language in executable behavior extends beyond smart home control. SayCan~\cite{saycan} extracts and leverages the knowledge within LLMs in physically-grounded tasks to select robot actions. ReAct~\cite{react} interleaves reasoning, actions, and knowledge based or environmental observations. Hierarchical structured prediction also has established precedence in Abstract Syntax Networks~\cite{asn}, that decode code generation tasks using syntax tree structures.

\name{} applies both these ideas to a runtime-defined smart home action space: its hierarchy exposes room, device, operation, and parameter decisions to a pretrained typed decision engine. Its contribution is its architecture, rather than hierarchical prediction or grounded action selection in isolation.

\subsection{Constrained Language-Model Inference}

Constrained decoding restricts generated outputs during inference. XGrammar~\cite{xgrammar} provides a flexible and efficient structure generation engine for LLMs. Synchromesh~\cite{synchromesh} allows enforcing syntactic and semantic constraints on the LLM output. SGLang~\cite{sglang} provides structured program execution and candidate selection.

\name{} is a domain-specific orchestration framework rather than a general language-model runtime. It explores the benefit of exposing smart home structure directly through hierarchical typed decisions.
\section{Conclusion}

\name{} reformulates smart home control as hierarchical typed decisions over a smart home's actions instead of open-ended device call generation. By restricting selections to valid candidates and abstaining when confidence is low, \name{} avoids unsupported actions and reduces incorrect executions. Its structured memory further enables personalized, multi-turn interactions without including conversation history in the model's prompt. To evaluate \name{}, we introduce 3 datasets extended from HomeBench~\cite{homebench}: Trek-Base, Trek-Robust and Trek-Memory. \name{} achieves 85.3\% accuracy on Trek-Robust versus 69.8\% for a finetuned Qwen3.5-4B baseline, while remaining competitive on Trek-Base requests. With memory, \name{} resolves 75.6\% of the memory requests, compared with 42.5\% for the finetuned baseline. These results demonstrate the effectiveness of hierarchical typed decisions for local smart home control.

\bibliography{references}
\bibliographystyle{mlsys2025}

\section{Appendix}

\lstdefinestyle{prompt}{basicstyle=\ttfamily\footnotesize, breaklines=true, columns=fullflexible,
  frame=single, framesep=4pt, xleftmargin=4pt, xrightmargin=4pt, keepspaces=true, captionpos=b}

\subsection{Latency Results}
\label{app:latency}

Tables \ref{tab:latency_Trek-Base}, \ref{tab:latency_Trek-Robust} and \ref{tab:memory-latency} show the end-to-end latency of \name{} and its baselines on Trek-Base, Trek-Robust and Trek-Memory, respectively.

\begin{table}[h]
\centering
\footnotesize
\setlength{\tabcolsep}{3pt}
\begin{tabular*}{\columnwidth}{
@{\extracolsep{\fill}}lccc@{}}
\toprule
 & Median (s) & P90 (s) & \# Calls \\
\midrule
\name{} & 0.166 & 0.801 & 5.4 \\
\midrule
Qwen3.5-4B, prompted & 0.573 & 0.813 & 1.0 \\
Qwen3.5-4B, finetuned & 0.531 & 0.897 & 1.0 \\
VCU-LLM & \textbf{0.079} & \textbf{0.151} & 1.8 \\
Domux & 0.169 & 0.797 & 1.0 \\
Sasha & 1.803 & 4.542 & 2.9 \\
AdaHome & 1.295 & 2.636 & 4.2 \\
\bottomrule
\end{tabular*}
\caption{End-to-end latency on Trek-Base (500 Requests).}
\label{tab:latency_Trek-Base}
\end{table}

\begin{table}[h]
\centering
\footnotesize
\setlength{\tabcolsep}{3pt}
\begin{tabular*}{\columnwidth}{
@{\extracolsep{\fill}}lccc@{}}
\toprule
 & Median (s) & P90 (s) & \# Calls \\
\midrule
\name{} & 0.169 & 0.607 & 4.5 \\
\midrule
Qwen3.5-4B, prompted & 0.647 & 1.578 & 1.0 \\
Qwen3.5-4B, finetuned & 0.643 & 1.684 & 1.0 \\
VCU-LLM & \textbf{0.094} & \textbf{0.160} & 1.4 \\
Domux & 0.169 & 0.345 & 1.0 \\
Sasha & 1.691 & 4.427 & 2.3 \\
AdaHome & 1.335 & 2.887 & 3.5 \\
\bottomrule
\end{tabular*}
\caption{End-to-end latency on Trek-Robust (500 Requests).}
\label{tab:latency_Trek-Robust}
\end{table}

\begin{table}[h]
\centering
\setlength{\tabcolsep}{4pt}
\resizebox{\columnwidth}{!}{%
\begin{tabular}{@{}lccc@{}}
\toprule
 & Median (s) & P90 (s) & Calls \\
\midrule
\name{}, no memory & 0.172 & 0.836 & 5.7 \\
\name{} + memory & 0.215 & 1.070 & 7.5 \\
\midrule
Qwen3.5-4B, finetuned & 0.486 & 0.821 & 1.0 \\
Qwen3.5-4B, finetuned + history & 0.727 & 1.673 & 1.0 \\
\bottomrule
\end{tabular}%
}
\caption{Latency per turn on Trek-Memory (3 conversations, 575 Requests).}
\label{tab:memory-latency}
\end{table}

\subsection{\name{}'s Prompts}
\label{app:heartprompts}

\subsubsection{The Wrapper Around Every JevK5 Typed Decision}

\medskip
\begin{lstlisting}[style=prompt, label={lst:jev-system},
  caption={System Message}]
Apply the supplied criterion to the supplied evidence. Choose exactly one listed option. Respond with only its uppercase letter, with no explanation or reasoning.
\end{lstlisting}

\medskip
\begin{lstlisting}[style=prompt, label={lst:jev-user},
  caption={User Message}]
{"evidence": <request text>, "criterion": <question>, "options": [{"letter": "A", "description": "<key>: <option text>"}, ...]}
\end{lstlisting}

\subsubsection{Typed Decisions for Hierarchical Action and Argument Resolution}

The typed decisions used by \name{} are in Table \ref{tab:typed-questions}.

\begin{table*}[t]
\centering
\small
\setlength{\tabcolsep}{5pt}
\begin{tabular}{@{}p{0.30\textwidth}p{0.65\textwidth}@{}}
\toprule
\textbf{Decision} & \textbf{Question} \\
\midrule
Level 1 -- Room &
Which room does this command refer to? \\

Level 2 -- Device &
Which device in that room does this command control? \\

Level 3 -- Operation &
Which operation on that device does this command ask for? \\

\midrule
Enumerated parameter &
Which setting does the command ask for on this device? \\

Numeric parameter, first pass &
What value between \{lo\} and \{hi\} does the command ask for? \\

Numeric parameter, refinement &
Refine the value: which exact number between \{lo\} and \{hi\} does the command ask for? \\

\bottomrule
\end{tabular}
\caption{Typed decision questions used by \name{} for hierarchical action and argument resolution.}
\label{tab:typed-questions}
\end{table*}

\subsubsection{Typed Decisions for Memory Operations}

The typed decisions used by \name{}'s Memory Module are in Table \ref{tab:memory-decisions}.

\begin{table*}[t]
\centering
\footnotesize
\setlength{\tabcolsep}{4pt}
\renewcommand{\arraystretch}{1.15}
\begin{tabularx}{\textwidth}{
@{}
>{\raggedright\arraybackslash}p{0.17\textwidth}
>{\raggedright\arraybackslash}p{0.27\textwidth}
>{\raggedright\arraybackslash}X
@{}}
\toprule
\textbf{Decision} & \textbf{Question} & \textbf{Options} \\
\midrule

Command Type
& What is the user doing?
& \textbf{now}: asking the home to do something now;
  \textbf{remember}: telling the assistant something to
  remember for the future;
  \textbf{revise}: correcting the last action or asking
  the assistant to forget something it remembers. \\

\addlinespace

Kind of statement
& What does the user want the assistant to remember or change?
& \textbf{preference}: a lasting preference for how a device
  should usually be set;
  \textbf{alias}: a name or nickname for a device;
  \textbf{forget}: forgetting a remembered preference or name;
  \textbf{correction}: the previous action used the wrong
  device and this names the right one. \\

\addlinespace

Preference operation
& Which setting of the device is the user talking about?
& The device's \texttt{set\_*} operations. \\

\addlinespace

Preference situation
& Which words describe the situation in which the user
  wants this setting?
& Phrases from the message, plus
  ``no particular situation -- this is the usual setting''. \\

\addlinespace

Alias
& Which words are the name the user wants to use for
  the device?
& Phrases from the message. \\

\addlinespace

Correction
& Which words in the earlier request referred to the device?
& Phrases from the earlier request. \\

\addlinespace

``it''
& Which device does this command refer to?
& Recently used devices, plus
  ``a device that was not used recently''. \\

\bottomrule
\end{tabularx}
\caption{Typed decisions used by \name{}'s Memory Module.}
\label{tab:memory-decisions}
\end{table*}

\subsection{Adapting HomeBench Prompt}
\label{app:prompt}

GPT-5 rewrites user requests with the system prompt. The \code{{style}} placeholder is replaced by one of the three style blocks. The user message is \code{COMMAND} (N=$n$): $u$, where $u$ is the user request and $n$ is its number of gold actions.

\medskip
\begin{lstlisting}[style=prompt, label={lst:prompt-system},
  caption={System Prompt to GPT-5 to Adapt HomeBench}]
Rewrite a smart-home command the way a real person would actually say it out loud.

You are given the command, N (the number of separate actions it contains), and a style.

Always:
- Produce EXACTLY N clauses, one per action, in the SAME order, about the SAME room each.
- Join clauses with ". Also," -- every clause after the first begins with "Also,". If N is 1, output one clause with no "Also,".
- KEEP every explicit number exactly as given, and keep the direction of a relative change ("by 18 degrees cooler" stays "18 degrees cooler"). A number the sentence drops is a number nothing can recover, and when the answer key refuses because a value is out of range, dropping it makes the row unanswerable rather than hard.
- Sound like speech, not like a manual. Contractions and politeness are good.
- Output ONLY the rewritten command. No quotes, no explanation.

{style}
\end{lstlisting}

\medskip
\begin{lstlisting}[style=prompt, label={lst:prompt-explicit},
  caption={Style Prompt for Explicit Requests}]
Style: EXPLICIT. Name the device and the room and give the value, the way a
real person actually says it -- contractions, "can you", "please", not a template.
Example: "Set the brightness of the light in the kitchen to 20." -> "Can you set the kitchen light to 20?"
\end{lstlisting}

\medskip
\begin{lstlisting}[style=prompt, label={lst:prompt-implied},
  caption={Style Prompt for Implied Device Requests}]
Style: DEVICE IMPLIED. Name the ROOM and what the person wants, but never name
the appliance. Keep any number.
Example: "Set the brightness of the light in the kitchen to 20." -> "Make the kitchen pretty dim, around 20."
\end{lstlisting}

\medskip
\begin{lstlisting}[style=prompt, label={lst:prompt-goal},
  caption={Style Prompt for Goal Only Requests}]
Style: GOAL ONLY. Say how things feel or what the person wants changed. Do not
name the appliance. The room may be stated or left to context.
Example: "Turn off the light in the kitchen." -> "It is too bright in the kitchen."
\end{lstlisting}

\subsection{Trek-Robust Prompt}
\label{app:Trek-Robustprompt}

\medskip
\begin{lstlisting}[style=prompt, label={lst:prompt-hard-core},
  caption={Trek-Robust Generation Prompt for Counterfactual, Unseen Capability, Missing Information and Novel Identity}]
--- system ---
You are building v4hard, a harder test set for smart-home command understanding. You receive one
original HomeBench test row and a CATEGORY. You return a PATCH that changes the home around the
device(s) the instruction concerns, the SAME instruction, and the correct output for that
instruction in the patched home. Your output is the answer key, so derive it from the patched home
and show your derivation.

INPUT
- SOURCE_ID, CATEGORY, CASE (missing_information: "A"|"B"; unseen_capability: "B1"|"B2"; else null)
- INSTRUCTION: the user's words, possibly vague. Return it verbatim. Never rewrite it.
- PARAPHRASE: explicit restatement, for your understanding only. Do not return it.
- TARGET: the room(s) and device(s) the instruction concerns, with their full status block and
  method lines from the original home. Other rooms exist but are not shown; do not touch them.
- ORIGINAL_GOLD: HomeBench's answer for this instruction on the original home, one item per
  sub-request: "room.device.operation(arg)" or "error_input".

STEP 0 - source check. Decide whether ORIGINAL_GOLD actually follows from TARGET and the
instruction under the gold rules below. If it does not (for example the gold says error_input but
the device, operation and a valid value are all present), set "source_consistent": false with a
one-line reason and still produce your variants; the row will be reviewed, not used blindly.

THE PATCH (applied by a program to the original home)
{"room_rename": {"from": "<room>", "to": "<new_room>", "aliases": ["<the words the user uses for THIS room>"]} | null,
 "devices": {"<room>": {"<device_name>": <status block> | null, ...}},
 "method": [<method entry>, ...],
 "registry": {...} | null}
- devices: a listed device REPLACES the original device of that name in that room; null REMOVES it;
  a new name ADDS it. Status block = {"state": "...", "attributes": {attr: {"value":..,"lowest":..,
  "highest":..} | {"value":..,"options":[..]} | {"value":..}}} plus optional "aliases": [...],
  "description": "...", "capabilities": "unknown", "supported": [...].
- method: the COMPLETE list of method entries for every device you listed (they replace all existing
  entries for those devices). Entry = {"room_name","device_name","operation","parameters":[{"name","type"}]}
  plus optional "availability": "available"|"unavailable"|"unknown" and, for a new operation, a
  one-sentence "description". Unlisted devices keep their original method lines.
- registry (missing_information CASE A only): {"drop_method": "all" | ["room.device.operation", ...],
  "drop_attributes": ["room.device", ...], "stale": true|false}. Changes what is SHOWN, not what is
  true; with "stale": true the method list keeps the old lines and the device's "supported" field
  gives the live list.

REALISM RULES (violations make the row unusable)
R1 Never rename or re-signature an EXISTING operation (set_brightness stays set_brightness). A new
   operation must add a capability the device did not have (a timer, a schedule, an oscillation, a
   child lock, a target humidity, an eco mode, a night light ...), with a description a user could act on.
R2 Added devices are real device types with the attributes and operations that type has elsewhere
   in HomeBench (light, air_conditioner, heating, fan, curtain, blinds, humidifier, dehumidifiers,
   air_purifiers, aromatherapy, media_player, trash, water_heater, garage_door, vacuum_robot). Never
   a device named after an attribute or a verb.
R3 Aliases must be truthful names for the SAME room or device (synonyms, brand names, friendly
   names). Never give a room or device the name or alias of a different room or device, and never
   make the instruction refer to a different entity than it did originally.
R4 Identifiers you invent must vary: mix styles such as "node-7f2", "zb_0a91", "hue_lamp_3",
   "sensibo-ac", "rm_north_wing", "snug", "annex"; do not reuse one id across rows.
R5 Exactly one semantic change per variant. Keep every value you did not change byte-identical.
R6 The patched home must leave EXACTLY ONE correct action per sub-request. A new operation (B2) must
   not duplicate what an existing operation already achieves for this instruction (if set_temperature
   can already make the room "2 degrees warmer", do not add adjust_temperature; add a capability the
   instruction needs and no existing operation provides, or set cannot_support).
R7 Added or rebuilt devices follow TYPE_SCHEMAS (the attributes, ranges, options and operations that
   device type has in HomeBench). Do not invent ranges or modes for a known type.

RULES FOR THE GOLD
G1 gold is a JSON list with exactly one item per sub-request, in the order the instruction states
   them. Item = "room.device.operation(arg)" using the exact identifiers in the PATCHED home, or
   "error_input". Integers without units, enum values verbatim, no spaces.
G2 Derive each item from the patched home, in this order: room not present -> error_input; device
   not present, or the wording matches two devices -> error_input; device "capabilities": "unknown"
   -> error_input; operation not in the SHOWN method list for that device, or availability
   unavailable/unknown -> error_input; relative command ("by 20", "dimmer") = current value + delta
   (no current value shown -> error_input); value outside lowest..highest or not in options ->
   error_input (never clamp); else the action.
G3 labels: one per item: EXECUTE, UNKNOWN_DEVICE, AMBIGUOUS_TARGET, UNSUPPORTED_OPERATION,
   INSUFFICIENT_CONTEXT, INVALID_PARAMETER. For a capability hidden by registry (CASE A) use
   INSUFFICIENT_CONTEXT.
G4 derivations: one line per item naming the facts it rests on (device, method line, range, value).

CATEGORIES
- counterfactual: exactly ONE semantic change that makes at least one gold item differ from
  ORIGINAL_GOLD: remove the device (null); add the missing device (R2) when ORIGINAL_GOLD is
  error_input because it was absent; remove the needed operation from method; shrink the integer
  range so the requested value is just outside it; for a relative command, move the current value
  by 15-25 so the answer changes but stays in range; set the needed operation's "availability":
  "unknown"; or add a second device of the same type (rename the original to type_1, add type_2)
  when the instruction refers to the device by its type word. Return EXACTLY two variants:
  {"role": "A", "patch": null, ...} and {"role": "B", "patch": {...}, ...}.
- unseen_capability, CASE B1 (composition, familiar operations): a light with no brightness
  control (remove set_brightness and the brightness attribute), a read-only device with no method
  lines, a combined device (e.g. ceiling_fan_light carrying the fan's and the light's method lines,
  replacing both), or an integer attribute turned into options ["low","medium","high"] with the
  SAME operation name now taking an enum. CASE B2 (new signature): ADD a new operation per R1 with
  its parameter {name, type int|enum}; put its range/options in the device attributes; the
  instruction must be satisfiable by it, otherwise set cannot_support. Return EXACTLY one variant
  {"role": "test", ...}.
- missing_information, CASE A: no semantic change; use "registry" to hide the capability the
  instruction needs. Two golds: gold = what a reader of the shown text can justify (error_input,
  INSUFFICIENT_CONTEXT) and gold_if_capability_assumed = the action with the original capabilities.
  CASE B: "availability": "unknown" on the needed operation, or "capabilities": "unknown" on the
  device; gold = error_input (INSUFFICIENT_CONTEXT); gold_if_capability_assumed = null. Return
  EXACTLY one variant {"role": "test", ...}.
- novel_identity: keep state and capabilities; change identifiers per R3/R4: rename the target
  device to a manufacturer-style id with aliases/description containing the device word the
  instruction uses (null the old name, add the new one with its method lines); or room_rename to an
  uncommon name with the instruction's room word as its alias; or add a same-type twin with a
  distinct name. The gold uses the NEW identifiers. If the wording cannot single out one device,
  gold is error_input with AMBIGUOUS_TARGET; never pick the first. If ORIGINAL_GOLD is error_input
  because the device is absent, keep it absent (rename something else in the room) so the gold
  stays error_input. Return EXACTLY one variant {"role": "test", ...}.

OUTPUT: one JSON object only, no prose.
{"source_id": "...", "category": "...", "case": "A"|"B"|"B1"|"B2"|null,
 "source_consistent": true|false, "source_note": null|"<why>",
 "change_summary": "...",
 "variants": [
   {"role": "A"|"B"|"test",
    "patch": null | {"room_rename": ..., "devices": {...}, "method": [...], "registry": {...}},
    "gold": ["..."], "labels": ["..."],
    "gold_if_capability_assumed": ["..."]|null,
    "derivations": ["item 1: ...", "..."]}
 ],
 "instruction": "<verbatim>",
 "cannot_support": null|"<why>"}

--- user template ---
SOURCE_ID: {id}
CATEGORY: {category}   CASE: {case}
INSTRUCTION: {input}
PARAPHRASE: {paraphrase}
ORIGINAL_GOLD: {gold}
TARGET (room -> device -> status; then method lines): {target}
TYPE_SCHEMAS (HomeBench device types: attributes and operations): {schemas}
\end{lstlisting}

\medskip
\begin{lstlisting}[style=prompt, label={lst:prompt-hard-homes},
  caption={Trek-Robust Generation Prompt for the 20 New Homes Used by New Home}]
--- system ---
You design smart homes for a benchmark. Write ONE complete home in HomeBench's JSON structure.

STRUCTURE
{"home_status": {"<room_name>": {"room_name": "<room_name>", "aliases": [...], "<device_name>": {"state": "...",
   "aliases": [...], "description": "...", "attributes": {"<attr>": {"value": .., "lowest": .., "highest": ..} |
   {"value": .., "options": [..]} | {"value": ..}}}, ...}, ...},
 "method": [{"room_name": "...", "device_name": "...", "operation": "...", "parameters": [{"name": "...", "type": "int"|"str"}],
             "description": "<one sentence, REQUIRED for every operation of a novel device type>"}, ...]}

REQUIREMENTS
H1 6 to 10 rooms. At least two room names that HomeBench does not use (HomeBench's are master_bedroom, guest_bedroom,
   living_room, ding_room, study_room, kitchen, bathroom, foyer, corridor, balcony, garage, store_room); e.g. nursery,
   pantry, attic, laundry, sunroom, mudroom, loft, home_office, basement, porch. Room aliases hold the words a user
   would say ("baby's room", "the office").
H2 20 to 30 devices in total. In most rooms, two or three devices of the SAME type with distinct names and truthful
   aliases (e.g. "ceiling_light" and "desk_lamp", both type light; aliases ["main light"] and ["reading lamp"]).
   The device NAME is what an action string must use; give every device "aliases" (1-3) and a "description".
H3 At least three device types absent from HomeBench (HomeBench has light, air_conditioner, heating, fan, curtain,
   blinds, humidifier, dehumidifiers, air_purifiers, aromatherapy, media_player, trash, water_heater, garage_door,
   vacuum_robot). Examples: doorbell, sprinkler, oven, ev_charger, smart_lock, pet_feeder, robot_mower, towel_rail,
   shower, dishwasher. Each gets 2-4 operations, preferring FAMILIAR operation names (turn_on, turn_off, set_mode,
   set_temperature, set_level, open, close, start, stop, lock, unlock) and a one-sentence description per operation.
H4 Familiar device types follow TYPE_SCHEMAS exactly (attribute names, ranges, options, operations).
H5 Every device in method exists in home_status and vice versa; every attribute an operation sets exists with a range
   or options; every parameter type is "int" or "str". Realistic current values.
H6 No two devices in one room share a name; aliases never repeat another device's name or alias in the same room.
H7 Output the JSON object only.

--- user template ---
HOME_NUMBER: {n}
THEME: {theme}
TYPE_SCHEMAS: {schemas}

{theme} cycles through:
  family house with a nursery and a home office
  city apartment with a loft and a balcony garden
  farmhouse with a mudroom, pantry and porch
  two-storey townhouse with a basement workshop
  retirement bungalow with a sunroom and laundry
  student share house with four bedrooms
  lakeside cabin with an attic and a boathouse
  duplex with a granny flat and a double garage
  studio with a mezzanine and a terrace
  villa with a gym, cellar and pool house

{schemas} is:
{"air_conditioner":{"attributes":{"temperature":{"range":[16,30]},"mode":{"options":["cool","heat","fan_only","dry"]},"fan_speed":{"options":["auto","low","medium","high"]},"swing":{"options":["auto","up","middle","down"]}},"operations":{"turn_on":[],"turn_off":[],"set_temperature":["temperature:int"],"set_mode":["mode:str"],"set_fan_speed":["fan_speed:str"],"set_swing":["swing:str"]}},"air_purifiers":{"attributes":{"mode":{"options":["auto","sleep"]},"fan_speed":{"options":["auto","low","medium","high"]}},"operations":{"turn_on":[],"turn_off":[],"set_mode":["mode:str"],"set_fan_speed":["fan_speed:str"]}},"aromatherapy":{"attributes":{"interval":{"range":[10,60]},"intensity":{"range":[0,100]}},"operations":{"turn_on":[],"turn_off":[],"set_interval":["interval:int"],"set_intensity":["intensity:int"]}},"blinds":{"attributes":{},"operations":{"open":[],"close":[]}},"curtain":{"attributes":{"degree":{"range":[0,100]}},"operations":{"open":[],"close":[],"set_degree":["degree:int"]}},"dehumidifiers":{"attributes":{"mode":{"options":["auto","sleep"]},"intensity":{"range":[0,100]},"tank":{"range":[0,100]}},"operations":{"turn_on":[],"turn_off":[],"set_mode":["mode:str"],"set_intensity":["intensity:int"]}},"fan":{"attributes":{"speed":{"options":["auto","low","medium","high"]},"swing":{"options":["auto","up","middle","down"]}},"operations":{"turn_on":[],"turn_off":[],"set_speed":["speed:str"],"set_swing":["swing:str"]}},"garage_door":{"attributes":{},"operations":{"open":[],"close":[]}},"heating":{"attributes":{"fan_speed":{"options":["auto","low","medium","high"]},"mode":{"options":["heat","fan_only"]},"temperature":{"range":[30,100]}},"operations":{"turn_on":[],"turn_off":[],"set_fan_speed":["fan_speed:str"],"set_mode":["mode:str"],"set_temperature":["temperature:int"]}},"humidifier":{"attributes":{"intensity":{"range":[0,100]},"mode":{"options":["auto","sleep"]},"tank":{"range":[0,100]}},"operations":{"turn_on":[],"turn_off":[],"set_intensity":["intensity:int"],"set_mode":["mode:str"]}},"light":{"attributes":{"brightness":{"range":[0,100]},"color":{"type":"list"}},"operations":{"turn_on":[],"turn_off":[],"set_brightness":["brightness:int"],"set_color":["color:typing.Tuple[int, int, int]"]}},"media_player":{"attributes":{"volume":{"range":[0,100]}},"operations":{"play":[],"pause":[],"stop":[],"set_volume":["volume:int"],"set_song":["song:str"],"set_artist":["artist:str"],"set_style":["style:str"]}},"trash":{"attributes":{},"operations":{"pack":[]}},"vacuum_robot":{"attributes":{},"operations":{"set_mode":["mode:str"],"set_cleaning_area":["area:str"]}},"water_heater":{"attributes":{"temperature":{"range":[30,100]},"mode":{"options":["heating","eco"]}},"operations":{"turn_on":[],"turn_off":[],"set_temperature":["temperature:int"],"set_mode":["mode:str"]}}}
\end{lstlisting}

\medskip
\begin{lstlisting}[style=prompt, label={lst:prompt-hard-newhome},
  caption={Trek-Robust Generation Prompt for New Home}]
--- system ---
You are building v4hard-newhomes. You receive a NEW home (one no model was trained on) and a user
instruction written for a different home. Produce the SAME request in the SAME words, re-targeted to this home, and
the correct output.

RE-TARGETING (keep everything about the sentence except the room/device words)
T1 Change only the room word and/or the device word so the instruction refers to a device in THIS home that supports
   the requested operation. Keep the register, politeness, vagueness, values and sentence shape. If the original says
   "Can you dim the balcony light by 43 percent, please?" and this home has a nursery with a ceiling_light, the new
   instruction is "Can you dim the nursery ceiling light by 43 percent, please?". Use the device's name or one of its
   aliases in natural words (underscores -> spaces). For a multi-request instruction, re-target each sub-request
   independently.
T2 CASE execute: every sub-request must resolve to exactly one device that supports the operation with a valid value.
   Prefer rooms and device types that HomeBench does not have, and devices that have same-type siblings in the room
   (then the wording must pick one by name or alias).
   CASE refuse: make the FIRST sub-request unanswerable in one of these ways, chosen to fit the sentence: the device
   type word alone where the room holds two of that type (AMBIGUOUS_TARGET); a device this room does not have
   (UNKNOWN_DEVICE); an operation this device does not support (UNSUPPORTED_OPERATION); a value outside the range
   (INVALID_PARAMETER). Other sub-requests, if any, should execute.
T3 Never invent a device, room or operation that is not in HOME.

GOLD RULES
G1 Resolve against HOME: room by name or alias (none -> UNKNOWN_DEVICE; several -> AMBIGUOUS_TARGET); device by name or
   alias first, else by type word (none -> UNKNOWN_DEVICE; two or more -> AMBIGUOUS_TARGET, never pick the first);
   operation in method (else UNSUPPORTED_OPERATION); relative command = current value + delta; value within
   lowest..highest / options (else INVALID_PARAMETER); else EXECUTE.
G2 gold: one item per sub-request in order, "room.device.operation(arg)" with THIS home's exact identifiers, or
   "error_input". labels: one per item. derivations: one line per item citing the facts used.
G3 novel_output: true if any gold item uses a device type or operation name absent from HomeBench's vocabulary.

OUTPUT (JSON only)
{"source_id": "...", "case": "execute"|"refuse", "instruction_original": "<verbatim input>", "instruction": "<re-targeted>",
 "edits": "<which words changed>", "intents": [...], "gold": ["..."], "labels": ["..."], "derivations": ["..."],
 "novel_output": true|false, "note": null|"<anything odd>"}

--- user template ---
SOURCE_ID: {id}
CASE: {case}
INSTRUCTION (original home): {input}
PARAPHRASE (original home, for understanding only): {paraphrase}
HOME: {home}
HOME_INVENTORY: {inventory}
\end{lstlisting}

\medskip
\begin{lstlisting}[style=prompt, label={lst:prompt-hard-newvocab},
  caption={Trek-Robust Generation Prompt for New Vocabulary}]
--- system ---
You are building v4hard-vocab: the same home and the same instruction, but the device API is written in
another ecosystem's grammar. You rewrite the method lines of every device in the TARGET room(s) into the assigned
GRAMMAR, give a mapping back to the canonical operations, and write the gold in the new grammar.

GRAMMARS
- ha:     Home Assistant service style, e.g. "light.turn_on(entity_id=guest_bedroom.light)",
          "light.turn_on(entity_id=guest_bedroom.light,brightness_pct=40)", "climate.set_temperature(entity_id=...,temperature=24)"
- vendor: vendor-prefixed camelCase methods, e.g. "hue.guest_bedroom.light.setLevel(pct=40)", "daikin.guest_bedroom.ac.setMode(mode=cool)"
- rest:   REST calls on one line, e.g. "PUT /rooms/guest_bedroom/devices/light/brightness {value:40}", "POST /rooms/guest_bedroom/devices/light/on"
- renamed: canonical shape with synonyms, e.g. "guest_bedroom.light.set_level(40)", "guest_bedroom.light.power_on()"

RULES
V1 Rewrite EVERY method line of every device in the TARGET room(s), each as {"room_name", "device_name", "operation"
   (canonical, unchanged), "parameters" (unchanged), "render": "<the line exactly as the registry shows it, with the
   parameter placeholder written as the grammar writes it>", "template": "<the exact output item with {value} where the
   argument goes, or no placeholder for no-argument operations>", "description": "<one sentence>"}.
   Example (vendor): render "hue.guest_bedroom.light.setLevel(pct:int)", template "hue.guest_bedroom.light.setLevel(pct={value})".
V2 The template, with the value filled in, IS the only accepted output string for that action. No spaces unless the
   grammar needs them inside braces; integers without units; enum values verbatim.
V3 Keep devices, state and capabilities unchanged. Only the grammar changes.
V4 gold: one item per sub-request in order, written with the templates, or "error_input" for refusals (unchanged
   token). It must equal ORIGINAL_GOLD mapped through your templates; if ORIGINAL_GOLD does not follow from the home,
   set source_consistent false and explain.
V5 All devices in the home that are NOT in the target rooms keep their canonical lines (you do not list them).

OUTPUT (JSON only)
{"source_id": "...", "grammar": "...", "instruction": "<verbatim>", "source_consistent": true|false, "source_note": null|"...",
 "method": [...], "gold": ["..."], "labels": ["..."], "derivations": ["..."]}

--- user template ---
SOURCE_ID: {id}
GRAMMAR: {grammar}
INSTRUCTION: {input}
PARAPHRASE: {paraphrase}
ORIGINAL_GOLD (canonical grammar): {gold}
TARGET (room -> device -> status; then canonical method lines): {target}
HOME_INVENTORY: {inventory}
\end{lstlisting}

\subsection{Trek-Memory Prompt}
\label{app:Trek-Memoryprompt}

\medskip
\begin{lstlisting}[style=prompt, label={lst:prompt-memory-system},
  caption={Trek-Memory Rewording Prompt}]
You reword sentences a person says to their smart-home assistant, so a test set does not sound templated.
You receive the inserted turns of one home's conversation, in order, as JSON. For each turn return ONE natural
rewording that a real person might say in that situation, keeping the meaning EXACTLY:
- Keep every number exactly as written, and do not add numbers.
- Keep nicknames exactly as written (they are quoted in "keep"). Do not explain them.
- Keep the room and device words listed in "keep". Do not mention any other room or device.
- A turn that says "it"/"its" must still say "it"/"its" and must not name the device.
- A request for "my usual" setting must still say "usual" and must not state a value.
- A statement of a preference, a nickname, a correction or a request to forget must stay that kind of statement.
- Keep the situation's key word (e.g. reading, movie, evening).
- Vary the wording across turns: questions, polite requests, casual speech, short and long forms.
Answer with JSON: {"rewrites": {"<id>": "<rewording>", ...}} covering every id.
\end{lstlisting}

\subsection{Baselines Prompt}
\label{app:baselineprompt}

\medskip
\begin{lstlisting}[style=prompt, label={lst:prompt-qwen-prompt},
  caption={Qwen3.5-4B, Prompted and Finetuned: HomeBench's Prompt (single user message; placeholders filled per request)}]
{system.txt}<home_state>
  The following provides the status of all devices in each room of the current household, the adjustable attributes of each device, and the threshold values for adjustable attributes:{home_state}
</home_state>
<device_method>
     The following provides the methods to control each device in the current household:{device_method}
</device_method>
{example.txt}-------------------------------
Here are the user instructions you need to reply to.
<User instructions:> 
{instruction}
<Machine instructions:>

\end{lstlisting}

\medskip
\begin{lstlisting}[style=prompt, label={lst:prompt-qwen-system},
  caption={Qwen3.5-4B: HomeBench's \texttt{system.txt}, inserted at \{system.txt\}}]
You are 'Al', a helpful AI Assistant that controls the devices in a house. Complete the following task as instructed or answer the following question with the information provided only. The current status of the device and the methods it possesses are provided below, please only use the methods provided. Output "error_input" when operating non-existent attributes and devices. Only output machine instructions and enclose them in {}.

\end{lstlisting}

\medskip
\begin{lstlisting}[style=prompt, label={lst:prompt-qwen-examples},
  caption={Qwen3.5-4B: HomeBench's \texttt{example.txt}, inserted at \{example.txt\}}]
Here are a few examples, your output format should be consistent with the results provided in the example:
<example1>
    <User instructions:> Lower the air conditioner temperature in the guest bedroom by 1 degree.
    <Machine instructions:> {guest_bedroom.air_conditione.set_temperature(26)}
</example1>
<example2>
    <User instructions:> Turn off the lights in the living room.
    <Machine instructions:> {living_room.light.turn_off()}
</example2>
<example3>
    <User instructions:> Turn on the light in the living room.
    <Machine instructions:> {error_input}  (If there is no device or if there is no attribute or method that can be operated on the device, please output as "error_input".)
</example3>
<example4>
    <User instructions:> Increase the intensity of the aromatherapy in the corridor by 4%, set the brightness of the lights in the garage to 40, adjust the fan speed of the air conditioner in the living room to low, and set the volume of the media player in the living room to 0.
    <Machine instructions:> {corridor.aromatherapy.set_intensity(30),error_input,error_input,living_room.media_player.set_volume(0),} 
</example4>

\end{lstlisting}

\medskip
\begin{lstlisting}[style=prompt, label={lst:prompt-vcu-system},
  caption={VCU-LLM System Message (followed by the retrieved devices; the request is the user message)}]
You are a smart home assistant. Given all available services, relevant devices, and relevant devices' status (could be None), analyze the provided instructions and generate the control plan to be executed.
\end{lstlisting}

\medskip
\begin{lstlisting}[style=prompt, label={lst:prompt-sasha-clarify},
  caption={Sasha, Step 1: Clarify}]

You are an AI that controls a smart home.
You will be provided with a set of home appliances and an instruction.
Respond, IN ONE WORD, whether it is possible to successfully carry out the instruction using the following devices.

Home appliances: {devices}

Instruction : {user_command}


{format_instructions}

\end{lstlisting}

\medskip
\begin{lstlisting}[style=prompt, label={lst:prompt-sasha-clarify-format},
  caption={Sasha, Step 1: Clarify (output format)}]
The output should be formatted as a JSON instance that conforms to the JSON schema below.

As an example, for the schema {"properties": {"foo": {"title": "Foo", "description": "a list of strings", "type": "array", "items": {"type": "string"}}}, "required": ["foo"]}
the object {"foo": ["bar", "baz"]} is a well-formatted instance of the schema. The object {"properties": {"foo": ["bar", "baz"]}} is not well-formatted.

Here is the output schema:
```
{"properties": {"response": {"title": "Response", "description": "Answer by YES or NO if the task is possible or the query can be answered", "type": "string"}, "explanation": {"title": "Explanation", "description": "Return an explanation of the response to the user", "type": "string"}}, "required": ["response", "explanation"]}
```
\end{lstlisting}

\medskip
\begin{lstlisting}[style=prompt, label={lst:prompt-sasha-filter},
  caption={Sasha, Step 2: Filter Devices}]

You are an AI that controls a smart home.
You will be provided with a set of home appliances and an instruction.
Extract the names of the devices that are relevant to execute the instruction

APPLIANCE LIST: {devices}

Instruction : {user_command}

Respond by returning a subset of the devices under APPLIANCE LIST
{format_instructions}

\end{lstlisting}

\medskip
\begin{lstlisting}[style=prompt, label={lst:prompt-sasha-filter-format},
  caption={Sasha, Step 2: Filter Devices (output format)}]
The output should be formatted as a JSON instance that conforms to the JSON schema below.

As an example, for the schema {"properties": {"foo": {"title": "Foo", "description": "a list of strings", "type": "array", "items": {"type": "string"}}}, "required": ["foo"]}
the object {"foo": ["bar", "baz"]} is a well-formatted instance of the schema. The object {"properties": {"foo": ["bar", "baz"]}} is not well-formatted.

Here is the output schema:
```
{"properties": {"devices": {"title": "Devices", "description": "A list that contains the devices that can potentially accomplish the task", "type": "array", "items": {"type": "string"}}}, "required": ["devices"]}
```
\end{lstlisting}

\medskip
\begin{lstlisting}[style=prompt, label={lst:prompt-sasha-plan},
  caption={Sasha, Step 3: Plan}]

You are an AI that controls a smart home.
You will be provided with a set of home appliances and an instruction.
Your job is to figure out how to use the set of home appliances to execute the instruction.

FILTERED DEVICES: {devices}

USER COMMAND: {user_command}

Return a json file that follows the structure of FILTERED DEVICES. Change the values of the
dictionary to best accomplish the USER COMMAND.

{format_instructions}

\end{lstlisting}

\medskip
\begin{lstlisting}[style=prompt, label={lst:prompt-sasha-plan-format},
  caption={Sasha, Step 3: Plan (output format)}]
The output should be formatted as a JSON instance that conforms to the JSON schema below.

As an example, for the schema {"properties": {"foo": {"title": "Foo", "description": "a list of strings", "type": "array", "items": {"type": "string"}}}, "required": ["foo"]}
the object {"foo": ["bar", "baz"]} is a well-formatted instance of the schema. The object {"properties": {"foo": ["bar", "baz"]}} is not well-formatted.

Here is the output schema:
```
{"properties": {"devices": {"title": "Devices", "description": "Return a diff json file that follows the structure of FILTERED DEVICES and specifies which values of the dictionary should be changed to best accomplish the USER COMMAND.", "type": "object"}}, "required": ["devices"]}
```
\end{lstlisting}

\medskip
\begin{lstlisting}[style=prompt, label={lst:prompt-adahome-classify},
  caption={AdaHome, Step 1: Classify the Request}]
You are the Intent Classifier for a smart-home assistant.
Input:
- User command: {user_command}
Classify the user command into exactly one category:
- DIRECT: explicit device-control instruction with a clearly specified action
and target device (e.g., "turn on the living room light")
- INDIRECT: implicit but inferable intent that can be resolved from context (e.g., "it is too hot")
- AMBIGUOUS: subjective, preference-dependent, or routine-like request that may
have multiple valid interpretations depending on the user (e.g., "I want to relax")
Respond ONLY with valid JSON.
The "category" key must strictly be one of the following: DIRECT, INDIRECT, or AMBIGUOUS.
Example:
{{"category": "DIRECT"}}
\end{lstlisting}

\medskip
\begin{lstlisting}[style=prompt, label={lst:prompt-adahome-plan-indirect},
  caption={AdaHome, Step 2a: Plan for Indirect or Ambiguous Requests}]
You are the Reasoning Planner for a smart-home assistant.
Input:
- Device states: {devices}
- User command: {user_command}
Infer the necessary device state changes based on the user command.
Think step by step, but keep only a minimal draft with at most 8 words.
Respond EXACTLY in the following format:
Draft: <concise reasoning draft>
Action: <valid JSON object with state changes>
Use only the device names provided in the device states.
\end{lstlisting}

\medskip
\begin{lstlisting}[style=prompt, label={lst:prompt-adahome-plan-direct},
  caption={AdaHome, Step 2b: Plan for Direct Requests}]
You are the Direct Planner for a smart-home assistant.
Input:
- Device states: {devices}
- User command: {user_command}
Infer the necessary device state changes based on the user command.
Respond EXACTLY in the following format:
Action: <valid JSON object with state changes>
Use only the device names provided in the device states.
\end{lstlisting}

\medskip
\begin{lstlisting}[style=prompt, label={lst:prompt-adahome-reprompt},
  caption={AdaHome, Step 3: Re-prompt When the Plan Has an Error}]

Your previous answer was:
{previous}
These state changes are not valid for this home: {errors}
Answer again in the same format, using only the device names, settings and values given in the device states.
\end{lstlisting}

\subsection{Table~\ref{tab:model-outputs} Prompt}
\label{app:tableprompt}

\medskip
\begin{lstlisting}[style=prompt, label={lst:prompt-table1},
  caption={Table~\ref{tab:model-outputs} Prompt to Qwen2.5-1.5B/3B/7B-Instruct (first request shown)}]
You control a smart home. These are the ONLY devices and the actions each supports:
balcony.light: turn_on(), turn_off(), set_brightness(0-100)
living_room.light: turn_on(), turn_off(), set_brightness(0-100)
living_room.thermostat: set_temperature(16-30)   # degrees Celsius
front_door.lock: lock(), unlock()

Current device state:
balcony.light = on, brightness 80
living_room.light = off
living_room.thermostat = 21   # degrees Celsius
front_door.lock = locked

User request: "dim the balcony light to 50%"

Output only the device action call(s) needed to satisfy the request, written as room.device.operation(argument). If several are needed, separate them with commas. Output just the call(s), nothing else.
\end{lstlisting}

\end{document}